\documentclass[american,english]{article}
\usepackage[T1]{fontenc}
\usepackage[latin9]{inputenc}
\usepackage{color}
\usepackage{babel}
\usepackage{array}
\usepackage{multirow}
\usepackage{algorithm2e}
\usepackage{varwidth}
\usepackage{amsmath}
\usepackage{graphicx}
\usepackage{geometry}
\PassOptionsToPackage{normalem}{ulem}
\usepackage{ulem}
\usepackage[pdfusetitle,
 bookmarks=false,
 breaklinks=false,pdfborder={0 0 1},backref=false,colorlinks=true]
 {hyperref}

\makeatletter

\providecommand{\tabularnewline}{\\}

\usepackage{cite}
\usepackage{amsmath,amssymb,amsfonts}
\usepackage{algorithmic}
\usepackage{graphicx}
\usepackage{textcomp}
\usepackage{xcolor}
\usepackage{balance}

\usepackage{colortbl} 
\definecolor{header_color}{rgb}{0.74,0.88,0.91}
\definecolor{even_color}{rgb}{0.9,0.9,0.9}
\definecolor{subheader_color}{rgb}{0.85,0.93,0.95}
\definecolor{childheader_color}{rgb}{1.0,0.93,0.87}

\ifdefined\showcaptionsetup
 \PassOptionsToPackage{caption=false}{subfig}
\fi
\usepackage{subfig}
\makeatother

\begin{document}
\title{Predicting Symptoms of Amotivation and Anhedonia among University
Students with a Novel Oversampling Method}
\author{Dang Nguyen\textsuperscript{1}\thanks{Corresponding author: Dang Nguyen (d.nguyen@deakin.edu.au)},
Bao Duong\textsuperscript{1}, Arun Kumar\textsuperscript{1}, Dat
Phan-Trong\textsuperscript{1}, Julian Berk\textsuperscript{1},\\
Taylor Braund\textsuperscript{2}, Kien Do\textsuperscript{1}, Debopriyo
Bal\textsuperscript{2}, Wu Yi Zheng\textsuperscript{2}, Leonard
Hoon\textsuperscript{1},\\
Jill Newby\textsuperscript{2}, Helen Christensen\textsuperscript{2},
Svetha Venkatesh\textsuperscript{1}, Alexis Whitton\textsuperscript{2},
Sunil Gupta\textsuperscript{1}\\
\textsuperscript{1}\textit{Applied Artificial Intelligence Initiative
(A\textsuperscript{2}I\textsuperscript{2}), Deakin University, Geelong,
VIC, Australia}\\
\textsuperscript{2}\textit{Black Dog Institute, University of New
South Wales, Sydney, NSW, Australia}}
\maketitle
\begin{abstract}
University students experience disproportionately high rates of common
mental health conditions, such as depression, which can impair learning,
social functioning, and overall well-being. Within this context, symptoms
of amotivation (i.e. loss of motivational drive) and anhedonia (i.e.
diminished interest or pleasure) are particularly debilitating, yet
they frequently go undetected. Developing new approaches to identify
students with prominent amotivation and anhedonia could enable earlier
and more targeted intervention.

Machine learning (ML) methods have increasingly been used to classify
individuals according to symptom severity. However, these ML models
often suffer from class imbalance, where the majority of cases fall
in the low-symptom group and relatively few in the high-symptom group.
This imbalance can reduce model accuracy and bias predictions. To
address this, studies commonly employ the popular oversampling strategy
SMOTE\foreignlanguage{american}{. However, SMOTE has a notable limitation:
it may generate invalid values for nominal variables.}

\selectlanguage{american}%
In this paper, we introduce a novel and effective oversampling method
that addresses this shortcoming. Our approach leverages a predictive
model to generate nominal variables, rather than interpolating them.
We validate our method on a large-scale GPS location dataset collected
from university students and demonstrate that it is significantly
better than existing oversampling approaches in predicting elevated
symptoms of amotivation and anhedonia.
\end{abstract}

\section{Introduction\label{sec:Introduction}}

University students experience high rates of depression. Although
evidence supports the effectiveness of a range of treatments, the
presence of prominent symptoms of \textit{amotivation} (loss of motivational
drive and difficulty initiating or sustaining activities) and \textit{anhedonia}
(reduced capacity to experience interest or pleasure) has been linked
to poorer treatment responses and worse long-term prognosis \cite{whitton2023distinct}.
Despite the increased illness burden associated with these symptoms
\foreignlanguage{american}{\cite{breslau2009impact,ducasse2018anhedonia,symonds2019development}},
they are often under-recognised, particular in young people \foreignlanguage{american}{\cite{schwan2021perceptions}.}
As a result, \textit{identifying new ways to detect and predict these
symptoms represents an important priority for early intervention}.

\selectlanguage{american}%
Machine learning (ML) methods have achieved significant successes
in many domains including education and healthcare \cite{habehh2021machine,hilbert2021machine,alanazi2022using}.
To predict amotivation and anhedonia (\textit{hereafter referred to
collectively as ``amotivation''}), prior research \cite{babic2017machine,orji2022machine,orji2023modeling}
simply applied popular ML models such as \textit{k-nearest neighbors}
(kNN), \textit{decision tree} (DT), \textit{support vector machine}
(SVM), and \textit{random forest} (RF) to student datasets containing
measures of amotivation. The data were mostly collected from student's
smartphones in time series format, which contained different types
of signals e.g. accelerometer, gyroscope, step count, and GPS location.
However, predicting amotivation is a challenging task for ML classifiers
because the training data are often imbalanced, where the proportion
of individuals with elevated symptoms of amotivation (i.e. \textit{minority
class}) is often much smaller than the proportion of individuals with
no or low amotivation symptoms (i.e. \textit{majority class}).

To rebalance the training set, recent studies e.g. \cite{orji2022machine,orji2023modeling}
leveraged a popular oversampling technique called SMOTE\foreignlanguage{english}{
\cite{chawla2002smote} to generate synthetic minority samples. SMOTE
generates additional synthetic minority samples where each sample
is generated by linearly combining two real minority samples. It is
showed to be comparable or better than other traditional methods like
AdaSyn \cite{he2008adasyn} and deep learning generative methods like
CTGAN \cite{Xu2019} and TVAE \cite{Xu2019,borisov2022deep}.}

\selectlanguage{english}%
However, \textit{SMOTE has a significant weakness}. As SMOTE simply
treats all variables as continuous variables, generated minority samples
may contain continuous values in \textit{nominal variables} (i.e.
the variables strictly contain integer values e.g. Treatment\_ID,
Trial\_ID, ...). SMOTE-NC \cite{fernandez2018smote} -- an improved
version of SMOTE, considers continuous and nominal variables separately.
It generates a nominal value by choosing the most frequent nominal
value among the k-nearest neighbors. Although it guarantees that the
generated nominal value is valid, this value is restricted to a small
range of nominal values in the minority class. As a result, it reduces
the diversity of the synthetic samples.

In this paper, we propose a novel method to generate a nominal variable.
We consider continuous variables as features and train a predictive
model to predict the nominal variable as a target variable. As there
is often no imbalance problem in the nominal variable, our predictive
model can effectively predict nominal values. Moreover, as our predictive
model is trained on the whole range of nominal values in the dataset,
it can generate diverse results, which is more generalizable than
SMOTE-NC. More importantly, in case imbalance happens in the nominal
variable, the performance of our predictive model is still as accurate
as SMOTE-NC because it tends to predict the most frequent nominal
value. We call our method \textbf{SMOTE-PRED}.

We apply our method SMOTE-PRED to a GPS location dataset collected
from Australia university students, which consists of 784 students
and 415,746 GPS locations. For each student, we have a sequence of
locations (indicated by latitude, longitude, and a timestamp) where
they visited during the trial period. Since the dataset is \textit{severely
imbalanced} (80\% ``low amotivation'' vs. 20\% ``high amotivation''),
we use SMOTE-PRED to rebalance it. Inspired by other researches using
GPS location data to predict mental health statuses such as stress
\cite{shvetcov2024passive}, schizophrenia \cite{barnett2018relapse,jongs2020framework},
and depression \cite{muller2021depression}, we extract patterns of
student's mobility as features to train ML classifiers to predict
student amotivation.

In summary, we make the following contributions:
\begin{enumerate}
\item \textbf{SMOTE-PRED -- an effective oversampling method:} We propose
an alternative strategy to generate nominal variables by predicting
them through a predictive model trained on continuous variables, rather
than interpolating their values directly.
\item \textbf{Real-world application:} We apply our proposed oversampling
method to \foreignlanguage{american}{a large-scale} GPS location dataset
collected from university students across Australia. Our approach
significantly outperforms existing oversampling methods, achieving
an AUC 0.67 compared to 0.64 from the best baseline and yielding a
8\% improvement over a standard classifier without oversampling.
\item \textbf{Mobility-amotivation link:} Our results reveal a meaningful
relationship between mobility patterns of students and their levels
of amotivation, offering new insight into behavioral indicators of
mental well-being.
\end{enumerate}
\selectlanguage{american}%
The remaining of the paper is organized as follows. In Section \ref{sec:Related-Works},
we summarize existing works for using GPS location data to predict
mental health problems. We also discuss current methods for oversampling,
including traditional ML methods and deep learning approaches. In
Section \ref{sec:Framework}, we focus on our main contributions,
where we describe our novel oversampling method SMOTE-PRED. In Section
\ref{sec:Experiments}, we first describe our GPS location dataset.
We then report and analyze the experimental results on our amotivation
predictions for university students. In Section \ref{sec:Conclusion},
we conclude our work and suggest future directions.\selectlanguage{english}%

\section{Related Works\label{sec:Related-Works}}

\subsection{Amotivation prediction}

Many methods have been proposed to predict student amotivation, which
can be categorized into two groups: traditional approaches and ML-based
approaches. Traditional methods use statistical tests such as Pearson's
correlation, t-test, and linear regression analysis to determine important
features for amotivation prediction \cite{vallerand1992academic,ilter2021relationship,schwan2021perceptions}.
In contrast, ML-based methods train modern ML classifiers to predict
student amotivation \cite{babic2017machine,orji2022machine,orji2023modeling}.

The training set of ML classifiers is often imbalanced, where the
number of students with ``low amotivation'' is significantly larger
than the number of students with ``high amotivation''. So far, existing
ML-based methods simply apply the well-known oversampling technique
SMOTE \cite{chawla2002smote} to rebalance the training set. However,
SMOTE has a significant weakness, which often produces continuous
values in nominal variables. In this paper, \textit{we propose a novel
oversampling method to address this problem}.

\subsection{GPS location data}

Recently, \textit{digital phenotyping} is considered as a new framework
to quantify health-related symptoms and behaviors \cite{palmius2016detecting,jongs2020framework,muller2021depression,raugh2020geolocation,shvetcov2024passive}.
It consists of passive sensor data collected from smartphones and
wearables. Its data sources can be varied from location (GPS), movement
(accelerometry), usage (gyroscope), and communication (SMS).

Many researches have demonstrated that using digital phenotyping may
be a promising method for detecting and monitoring mental health \cite{onnela2016harnessing,neary2018state,liang2019survey,bufano2023digital}.
Among different types of smartphone data, GPS location is the most
popular data type used in digital phenotyping studies \cite{beames2024use}.
GPS location data have been used to predict mental health statuses
such as stress \cite{shvetcov2024passive}, schizophrenia \cite{barnett2018relapse,jongs2020framework},
and depression \cite{muller2021depression}.

Nevertheless, working on GPS location data is a challenging task \cite{shvetcov2024passive}.
GPS location data often have inaccurate coordinates due to a high
dependency on many factors such as unblocked receivers and good signals.
Moreover, intermittent smartphone connectivity, caused by internet
or signal loss or by the phone being switched off, leads to data loss
and consequently a high rate of missing data. More importantly, although
we can extract many features from a GPS location dataset, most of
them are irrelevant. If we do not select features carefully, training
a ML classifier with small high-dimensional dataset can easily cause
the model over-fitting. In this paper, \textit{we introduce a comprehensive
pipeline from pre-processing data, extracting useful features, and
imputing missing data to addressing imbalanced data and training ML
classifiers}.

\subsection{Imbalanced classification}

\textit{Tabular classification} is the most well-known task in tabular
data. It predicts a label for the target variable (e.g. \textquotedblleft Diabetes\textquotedblright )
by using a set of features (e.g. \textquotedblleft BMI\textquotedblright{}
and \textquotedblleft HbA1c\textquotedblright ). While deep learning
methods are often dominant in other tasks, traditional ML classifiers
still outperform deep learning methods in tabular classification \cite{shwartz2022tabular,Grinsztajn2022}.

\textit{Imbalanced classification} often happens when one class in
the training set has a much larger numbers of samples than the others.
There are two main approaches to address the class-imbalance. Model-centric
approaches focus on modifying the objective functions in the ML classifiers
\cite{Cao2019} or re-weighting the minority classes \cite{Cui2019}.
A data-centric approach is oversampling, which generates more synthetic
minority samples. 

\textit{Oversampling} methods are well-known solutions for imbalanced
classification. Most existing methods based on SMOTE (Synthetic Minority
Oversampling Technique) \cite{chawla2002smote}, in which a new minority
sample is generated by linearly combining two real minority samples.
Several variants have been developed to address SMOTE weaknesses such
as outlier and noise \cite{batista2003balancing,batista2004study,han2005borderline,nguyen2011borderline}.
Other approaches are generative deep learning models such as CTGAN
\cite{Xu2019} and TVAE \cite{Xu2019,borisov2022deep}, which learn
the distribution of real minority samples via a generator or encoder
network. Recently, large language models (LLMs) are also adapted to
oversampling \cite{Yang2024,nguyen2025large}.

In mental heath domain, deep learning based oversampling methods often
do not perform well due to typically small size of the training set.
In contrast, SMOTE-based oversampling methods do not properly address
mixed data types (continuous vs. nominal). In this paper, \textit{we
propose a novel SMOTE-based oversampling method that overcomes this
weakness}.

\section{Framework\label{sec:Framework}}

We first describe our problem i.e. how to use oversampling for imbalanced
classification. We then introduce our proposed oversampling method
SMOTE-PRED.

\subsection{Oversampling for imbalanced classification}

Let ${\cal D}_{train}=\{x_{i},y_{i}\}_{i=1}^{N}$ be an \textit{imbalanced}
tabular dataset, where each row consists of a sample $x_{i}$ (with
$M$ predictor variables $\{X_{1},...,X_{M}\}$) and a label $y_{i}$
(i.e. a value of the target variable $Y$). Without loss of generality,
we assume that $\{X_{1},...,X_{M-1}\}$ are \textit{continuous} variables
while $X_{M}$ is a \textit{nominal} variable (i.e. $X_{M}$ contains
only integer numbers). In case $X_{M}$ is a \textit{categorical}
variable, we convert it to a nominal variable using an ordinal encoding.
We consider $Y$ as a binary variable i.e. $y_{i}\in\{0,1\}$. We
also call class $Y=0$ as \textit{majority} (or \textit{negative})
class while class $Y=1$ as \textit{minority} (or \textit{positive})
class. The sets of majority and minority samples are denoted as ${\cal D}_{major}$
and ${\cal D}_{minor}$. We have ${\cal D}_{train}={\cal D}_{major}\cup{\cal D}_{minor}$
and $\mid{\cal D}_{minor}\mid\ll\mid{\cal D}_{major}\mid$.

\begin{figure}[h]
\begin{centering}
\includegraphics[scale=0.5]{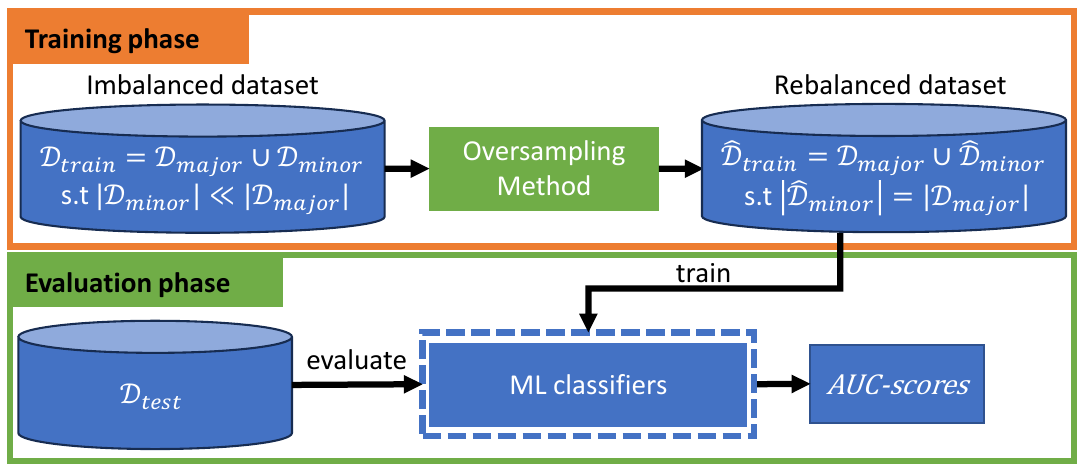}
\par\end{centering}
\caption{\label{fig:Training-and-evaluation}Training and evaluation phases
of an oversampling method. \textbf{Training:} the oversampling method
learns from the imbalanced dataset ${\cal D}_{train}$ to generate
synthetic minority samples $\hat{{\cal D}}_{minor}$ to construct
the rebalanced dataset $\hat{{\cal D}}_{train}$. \textbf{Evaluation:}
$\hat{{\cal D}}_{train}$ is used to train ML classifiers, and the
classifiers are evaluated on a held-out test set ${\cal D}_{test}$
to compute performance metrics (e.g. AUC-scores). \textit{A better
score implies a better oversampling method}.}
\end{figure}

An oversampling method aims to learn a data synthesizer from ${\cal D}_{train}$
to generate \textit{synthetic} minority samples $\hat{{\cal D}}_{minor}$
such that $\mid\hat{{\cal D}}_{minor}\mid=\mid{\cal D}_{major}\mid$.
Next, $\hat{{\cal D}}_{minor}$ is used to construct a \textit{rebalanced}
dataset $\hat{{\cal D}}_{train}={\cal D}_{major}\cup\hat{{\cal D}}_{minor}$.
Finally, the performance of an oversampling method is measured by
AUC-scores of ML classifiers trained on $\hat{{\cal D}}_{train}$
and tested on a held-out dataset ${\cal D}_{test}$. A better score
indicates a better oversampling method. Figure \ref{fig:Training-and-evaluation}
illustrates the training and evaluation phases for an oversampling
method.

\subsection{The proposed method}

We propose a SMOTE-based oversampling method (called \textbf{SMOTE-PRED}),
which is described in the following section.

\subsubsection{SMOTE-PRED a new way to re-correct nominal values}

Given a real minority sample $x_{i}$, SMOTE \cite{chawla2002smote}
generates a new synthetic minority sample $\hat{x}_{i}$ as follows:
\begin{equation}
\hat{x}_{i}=x_{i}+\lambda\times(x_{i}-x_{j}),\label{eq:SMOTE}
\end{equation}
where $x_{j}$ is another real minority sample and $\lambda\in(0,1)$
is a random number.

To compute the distance $(x_{i}-x_{j})$, SMOTE simply treats all
variables as continuous variables and applies an Euclidean distance
$d(x_{i},x_{j})$. As a result, the synthetic sample $\hat{x}_{i}$
has a continuous value for its nominal variable $X_{M}$, which causes
$\hat{x}_{i}$ unrealistic. For example, in healthcare domains, a
nominal variable can be \textit{treatment\_id}, which presents a type
of treatment. Therefore, it strictly contains only integers.

Our goal is to re-correct continuous values of $X_{M}$ generated
by SMOTE. First, we assume that the nominal variable $X_{M}$ is a
\textit{fake} target variable and its each possible value $v_{m}\in X_{M}$
is a class. We observe that the imbalanced problem often happens in
the real target variable $Y$ but not in the nominal variable $X_{M}$.
For example, the number of patients who receives each type of treatment
is often equal.

Next, from the original dataset ${\cal D}_{train}=\{x_{i},y_{i}\}_{i=1}^{N}$,
we construct a new dataset ${\cal D}'_{train}=\{x'_{i},v_{i,M}\}_{i=1}^{N}$,
where $x'_{i}$ is a subset of the original sample $x_{i}$ with only
continuous variables $\{X_{1},...,X_{M-1}\}$ and $v_{i,M}$ is a
value of the nominal variable $X_{M}$. Then, we train a predictive
model $f()$ on ${\cal D}'_{train}$, with the goal of predicting
the nominal variable $X_{M}$ based on continuous variables $\{X_{1},...,X_{M-1}\}$.

Finally, from the synthetic minority samples $\hat{{\cal D}}_{minor}$
generated by SMOTE, we extract continuous variables and use them as
input features for our trained predictive model $f()$. We then predict
integer values for $X_{M}$, and replace continuous values in $X_{M}$
generated by SMOTE with our predicted integers.

\begin{figure}[h]
\begin{centering}
\includegraphics[scale=0.4]{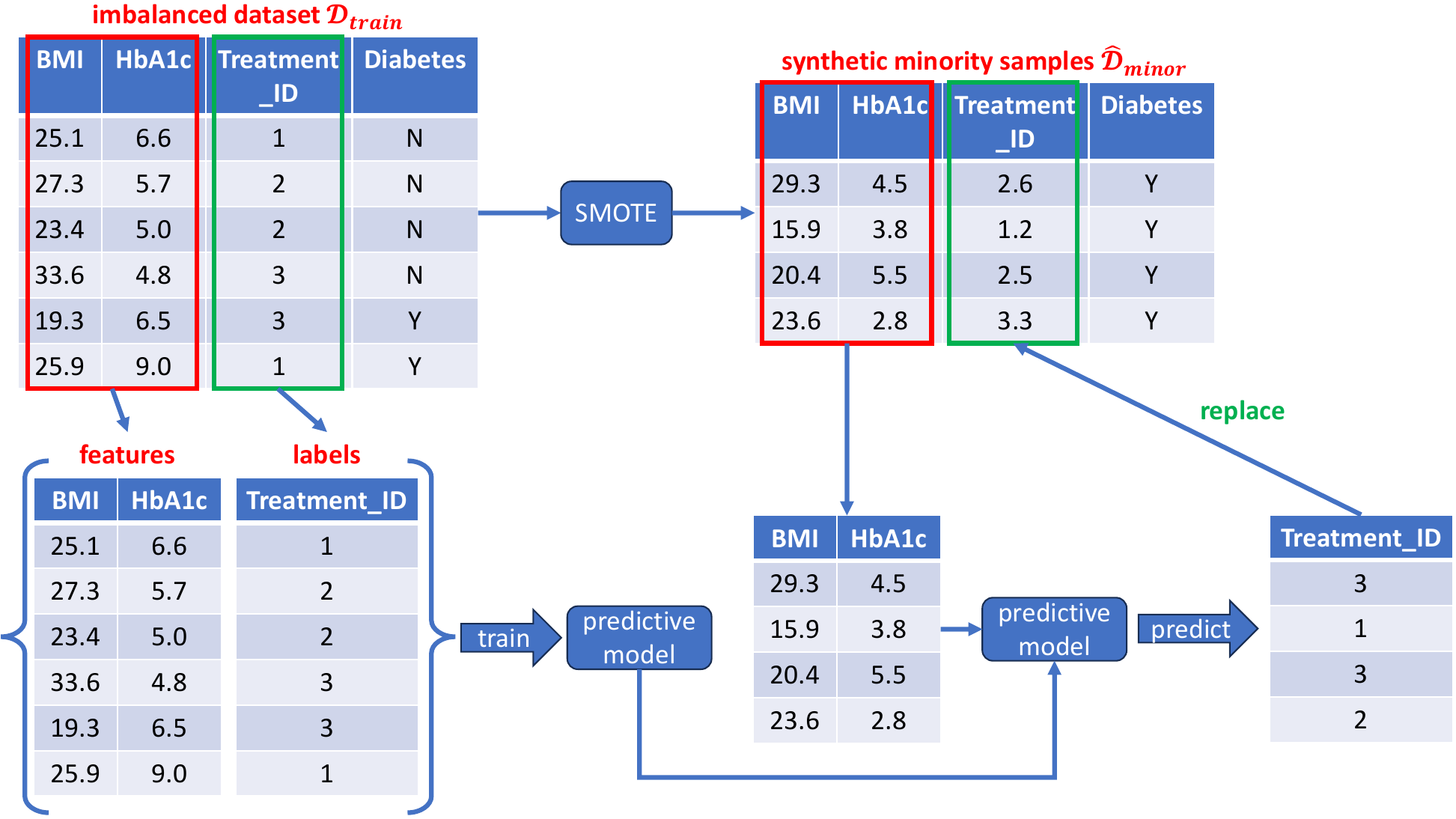}
\par\end{centering}
\caption{\label{fig:Our-PRE-component}Our SMOTE-PRED has three steps. First,
we apply SMOTE to the imbalanced dataset ${\cal D}_{train}$ to generate
synthetic minority samples $\hat{{\cal D}}_{minor}$. As SMOTE treats
all variables as continuous variables, it generates continuous values
for the nominal variable \textit{Treatment\_ID}. Second, from ${\cal D}_{train}$,
we extract continuous and nominal variables, and use them as features
and labels respectively to train a predictive model. Finally, we extract
continuous variables from $\hat{{\cal D}}_{minor}$, and use them
as input features for our trained predictive model to predict values
for the nominal variable \textit{Treatment\_ID}. We then replace generated
continuous values in \textit{Treatment\_ID} by SMOTE with our predicted
integers (i.e nominal values).}

\end{figure}

We illustrate three steps of SMOTE-PRED in Figure \ref{fig:Our-PRE-component}.

\subsubsection{Discussion}

Recall that given a real minority sample $x_{i}$, SMOTE-NC \cite{chawla2002smote}
generates a value for the nominal variable $X_{M}$ of the synthetic
minority $\hat{x}_{i}$ by choosing the most frequent nominal value
among k-nearest neighbors of $x_{i}$. However, as SMOTE-NC only focuses
on the minority class, the range of nominal values is limited. For
example, in Figure \ref{fig:Our-PRE-component}, the values of \textit{Treatment\_ID}
in the minority class (i.e. \textit{Diabetes=Y}) are limited to \{1,
3\}. Therefore, SMOTE-NC can only generate \textit{Treatment\_ID=1}
and \textit{Treatment\_ID=3} for synthetic minority samples. This
behavior reduces the diversity of the synthetic samples. In contrast,
since our SMOTE-PRED is trained on the full range of \textit{Treatment\_ID},
it can generate \textit{Treatment\_ID=2} for synthetic minority samples
as shown in Figure \ref{fig:Our-PRE-component}.

More importantly, although our SMOTE-PRED assumes the imbalanced problem
does not happen in the nominal variable $X_{M}$, in case it happens
our method still performs as well as SMOTE-NC. This is because our
predictive model will tend to predict the most frequent nominal value
in $X_{M}$.

\section{Experiments\label{sec:Experiments}}

\selectlanguage{american}%
In this section, we first describe our GPS location dataset. We then
explain how we pre-process data, extract features, and impute missing
data. Next, we present and discuss the result of our oversampling
method. Finally, we analyze our method under different configurations.

\subsection{GPS location data}

Our GPS location dataset was collected from the Vibe-Up study \cite{huckvale2023protocol,newby2025brief}
and approved by the University of New South Wales Human Research Ethics
Committee (approval number HC200466).

The dataset contains \foreignlanguage{english}{1,282} university students
with elevated symptoms of psychological distress (i.e. elevated symptoms
of depression, anxiety, and stress), of whom, \foreignlanguage{english}{784}
had usable GPS data that included 415,746 GPS locations. Data were
collected in the context of an adaptive clinical trial involving 12
sequential mini trials, all run between 2021-2023. Note that the GPS
locations were only recorded when the phone had been turned on or
a student had significantly moved. Moreover, each student has an amotivation
score $s\in[0,9]$, which was derived from a subset of items from
the Depression, Anxiety, and Stress Scales-21 instrument. Using cut-offs
established from a normative sample \cite{crawford2003depression},
we categorized the participants into those \textit{high on symptoms
of amotivation} ($s\geq5$) relative to those with \textit{no or minimal
symptoms of amotivation} ($s<5$).

For an illustrative purpose, we show some \textit{fake examples} of
our GPS location dataset in Table \ref{tab:Example-GPS-data}. Table
\ref{tab:Example-GPS-data}(a) shows students along with their visited
locations (in terms of \textit{latitude} and \textit{longitude}) during
the trial period (indicated by \textit{timestamp}). Table \ref{tab:Example-GPS-data}(b)
shows amotivation scores of students.

\begin{table}[h]
\caption{\label{tab:Example-GPS-data}Some \textbf{fake examples} of our GPS
location dataset for an illustrative purpose. Table (a) shows two
students along with their visited locations. Each student has a unique
ID and multiple GPS locations (in terms of latitude and longitude).
Each of them is associated with a timestamp. \textquotedblleft Horizontal
Accuracy\textquotedblright{} and\textquotedblleft Vertical Accuracy\textquotedblright{}
variables measure the accuracy range (in meters) of the recorded locations.
Table (b) shows the amotivation scores of these two students. Each
score $s$ is in a range of $[0,9]$, where $s<5$ is \textquotedblleft\textit{no
or minimal symptoms of amotivation}\textquotedblright{} whereas $s\protect\geq5$
is a \textquotedblleft\textit{high symptoms of amotivation}\textquotedblright .}

\begin{centering}
\subfloat[Students with GPS locations.]{
\centering{}%
\begin{tabular}{|l|l|r|r|r|r|}
\hline 
\textbf{Student} & \textbf{Timestamp} & \textbf{Latitude} & \textbf{Longitude} & \foreignlanguage{english}{\textbf{Horizontal Accuracy}} & \foreignlanguage{english}{\textbf{Vertical Accuracy}}\tabularnewline
\hline 
\hline 
abcxyz12 & 15/11/2021 07:35:00 & -30.8036 & 118.5869 & \foreignlanguage{english}{33} & \foreignlanguage{english}{24.5}\tabularnewline
\hline 
abcxyz12 & 15/11/2021 07:55:05 & -30.8879 & 118.5271 & \foreignlanguage{english}{20} & \foreignlanguage{english}{30}\tabularnewline
\hline 
abcxyz12 & 16/11/2021 08:36:39 & -30.9712 & 118.6020 & \foreignlanguage{english}{7.9} & \foreignlanguage{english}{23.9}\tabularnewline
\hline 
\foreignlanguage{english}{12abc456} & 19/11/2021 16:11:31 & -17.4219 & 142.9478 & \foreignlanguage{english}{15} & \foreignlanguage{english}{10}\tabularnewline
\hline 
\end{tabular}}
\par\end{centering}
\centering{}\subfloat[Amotivation scores.]{
\centering{}%
\begin{tabular}{|l|r|}
\hline 
\textbf{Student} & \textbf{Score}\tabularnewline
\hline 
\hline 
abcxyz12 & \foreignlanguage{english}{2}\tabularnewline
\hline 
\foreignlanguage{english}{12abc456} & \foreignlanguage{english}{6}\tabularnewline
\hline 
\end{tabular}}
\end{table}

\subsubsection{Data pre-processing}

Following \cite{raugh2020geolocation}, we pre-process our dataset
by filtering out inaccurate GPS locations. First, each geolocation
coordinate is paired with an accuracy range. This range indicated
by ``Horizontal Accuracy'' and ``Vertical Accuracy'' (refer Table
\ref{tab:Example-GPS-data}(a)) presents a range (in meters) of all
possible coordinates from which a single coordinate pair can be selected.
If this range is large, it means the coordinate pair ``Latitude''
and ``Longitude'' may not be recorded correctly. As a result, we
only select GPS locations whose accuracy range is less than 35 meters
as suggested in \cite{raugh2020geolocation}.

Second, two locations can be considered as duplicated if the change
in their coordinates is small \cite{palmius2016detecting,raugh2020geolocation}.
Following \cite{raugh2020geolocation}, we measure the change by calculating
Haversine distance between two locations $A=[lat_{A},long_{A}]$ and
$B=[lat_{B},long_{B}]$ as follows:
\[
a=\text{sin}(\frac{\bigtriangleup lat}{2})^{2}+\text{cos}(lat_{A})\times\text{cos}(lat_{B})\times\text{sin}(\frac{\bigtriangleup long}{2})^{2}
\]
\begin{align}
\text{Haversine}(A,B) & =2\text{arcsin}(\sqrt{a})\times R\label{eq:Haversine-dist}
\end{align}
where $\bigtriangleup lat=lat_{A}-lat_{B}$, $\bigtriangleup long=long_{A}-long_{B}$,
and $R=6371$ is Earth's radius (kilometers).

Given two locations A and B, we remove one of them if their distance
$\text{Haversine}(A,B)<500$ (meters) following \cite{raugh2020geolocation}.

We also excluded students with fewer than two days of GPS data, as
this duration was insufficient to capture their mobility patterns.

\begin{figure}[h]
\selectlanguage{english}%
\begin{centering}
\subfloat[]{\begin{centering}
\includegraphics[scale=0.3]{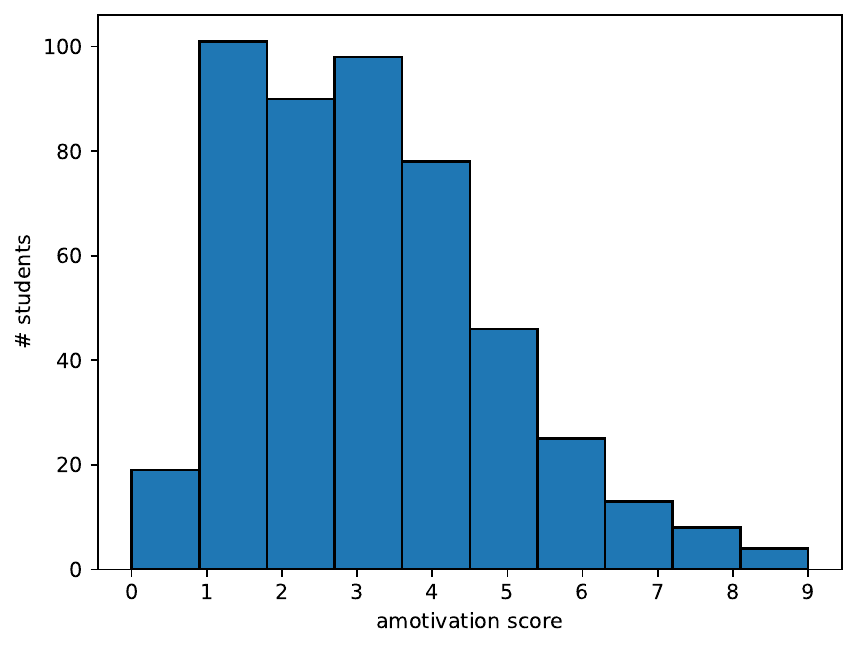}
\par\end{centering}
}\subfloat[]{\begin{centering}
\includegraphics[scale=0.3]{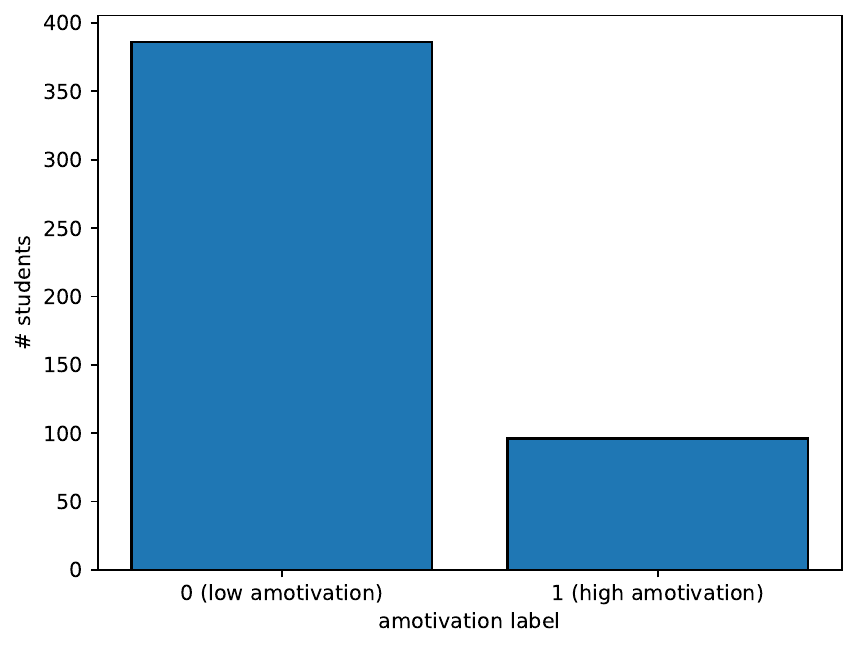}
\par\end{centering}
}
\par\end{centering}
\caption{\label{fig:Distributions-of-amotivation}Distributions of amotivation
scores (a) and labels (b). Among 482 students, 80\% of them are \textquotedblleft low
amotivation\textquotedblright{} (i.e. majority class) and 20\% are
\textquotedblleft high amotivation\textquotedblright{} (i.e. minority
class).}
\selectlanguage{english}%
\end{figure}

After filtering out inaccurate locations and removing duplicated locations,
our dataset contains 482 students and 2,211 GPS locations. We also
convert amotivation scores $s\in[0,9]$ to labels by following the
rule: if $s<5$, the label is ``low amotivation'' (\textit{majority
label}) otherwise the label is ``high amotivation'' (\textit{minority
label}). Figure \ref{fig:Distributions-of-amotivation} shows the
distributions of amotivation scores and amotivation labels in our
dataset. There are 386 students with ``low amotivation'' (80\%)
and 96 students with ``high amotivation'' (20\%).

\subsubsection{Feature extraction}

Although there are many different features that can be extracted from
a GPS location dataset \cite{barnett2018relapse,raugh2020geolocation,muller2021depression},
we derive and use two new features in our method and experiments.
They are (1) \textit{the number of visited locations} and (2) \textit{the
average distance traveled}. Both features are measured for each day.

First, we define a GPS location as $L_{i}=[t_{i},C_{i}]$, where $t_{i}$
is a timestamp and $C_{i}$ is a pair of latitude and longitude. We
then represent each day of a student as a sequence of location-time
points ordered by times as $\mathcal{L}=\{L_{1},...,L_{P}\}$.

\selectlanguage{english}%
\textbf{The number of visited locations }$n_{location}$\textbf{:}
We define this variable as the number of times the student moves to
a different location in a day, which is $n_{location}=\mid\mathcal{L}\mid=P$.
For example, from Table \ref{tab:Example-GPS-data}, the first student
``\foreignlanguage{american}{abcxyz12}'' visited two locations in
15/11/2021, therefore $n_{location}=2$ while they visited only one
place in 16/11/2021, therefore $n_{location}=1$.

\textbf{The average distance traveled $dist_{travel}$:} Given the
first location $C_{1}$, we define this variable as how far the student
travels in a day:
\begin{equation}
dist_{travel}=\frac{1}{P}\sum_{i=2}^{P}\mid\text{Haversine}(C_{1},C_{i})\mid,\label{eq:Average-dist}
\end{equation}
where $\mid\cdot\mid$ is an absolute value operator and $\text{Haversine}(C_{1},C_{i})$
is the Haversine distance computed in Equation (\ref{eq:Haversine-dist}).

For example, from Table \ref{tab:Example-GPS-data}, the first student
``\foreignlanguage{american}{abcxyz12}'' has $dist_{travel}=4.01$
in 15/11/2021 while has $dist_{travel}=0$ in 16/11/2021 as there
is only one location in this day.

Finally, we represent our GPS location data using the two new features
as shown in Table \ref{tab:Our-GPS-location}.

\begin{table}[h]
\caption{\label{tab:Our-GPS-location}Our \textbf{fake examples} in Table \ref{tab:Example-GPS-data}(a)
are represented with two features \textquotedblleft the number of
visited locations\textquotedblright{} $n_{location}$ and \textquotedblleft the
average distance traveled\textquotedblright{} $dist_{travel}$.}

\selectlanguage{american}%
\begin{centering}
\begin{tabular}{|l|l|r|r|}
\hline 
\textbf{Student} & \foreignlanguage{english}{\textbf{Date}} & \foreignlanguage{english}{$n_{location}$} & \foreignlanguage{english}{$dist_{travel}$}\tabularnewline
\hline 
\hline 
\multirow{2}{*}{abcxyz12} & 15/11/2021 & \foreignlanguage{english}{2} & \foreignlanguage{english}{4.01}\tabularnewline
\cline{2-4}
 & 16/11/2021 & \foreignlanguage{english}{1} & \foreignlanguage{english}{0}\tabularnewline
\hline 
\foreignlanguage{english}{12abc456} & 19/11/2021 & \foreignlanguage{english}{1} & \foreignlanguage{english}{0}\tabularnewline
\hline 
\end{tabular}
\par\end{centering}
\selectlanguage{english}%
\end{table}

\selectlanguage{american}%

\subsubsection{Missing data imputation}

\begin{figure*}
\selectlanguage{english}%
\begin{centering}
\subfloat[]{\begin{centering}
\includegraphics[scale=0.26]{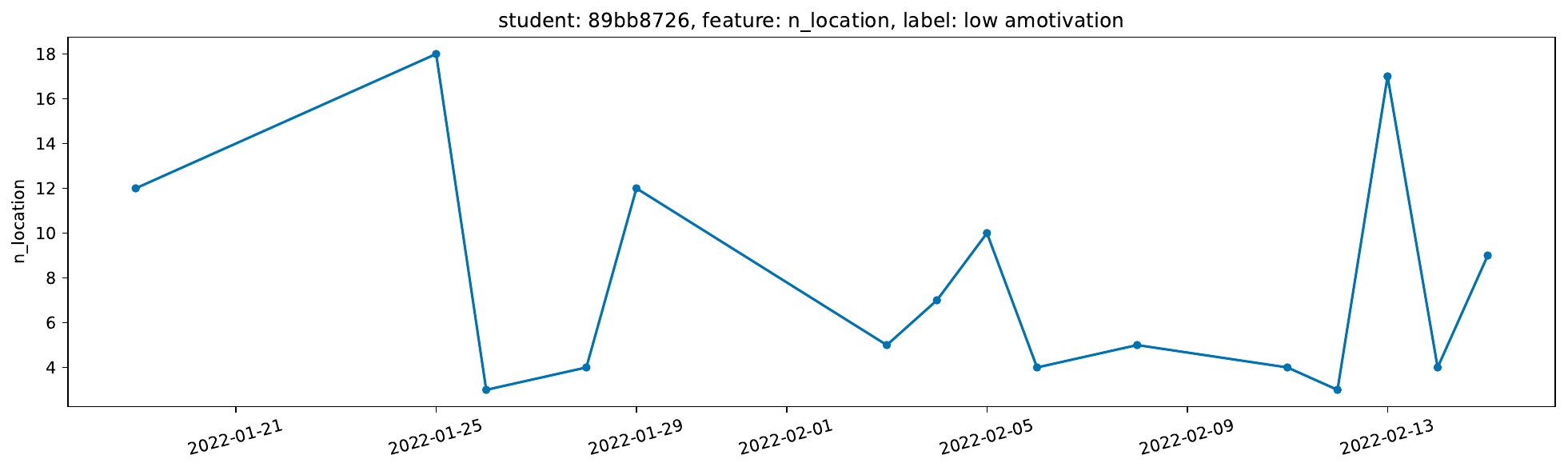}
\par\end{centering}
}\subfloat[]{\begin{centering}
\includegraphics[scale=0.26]{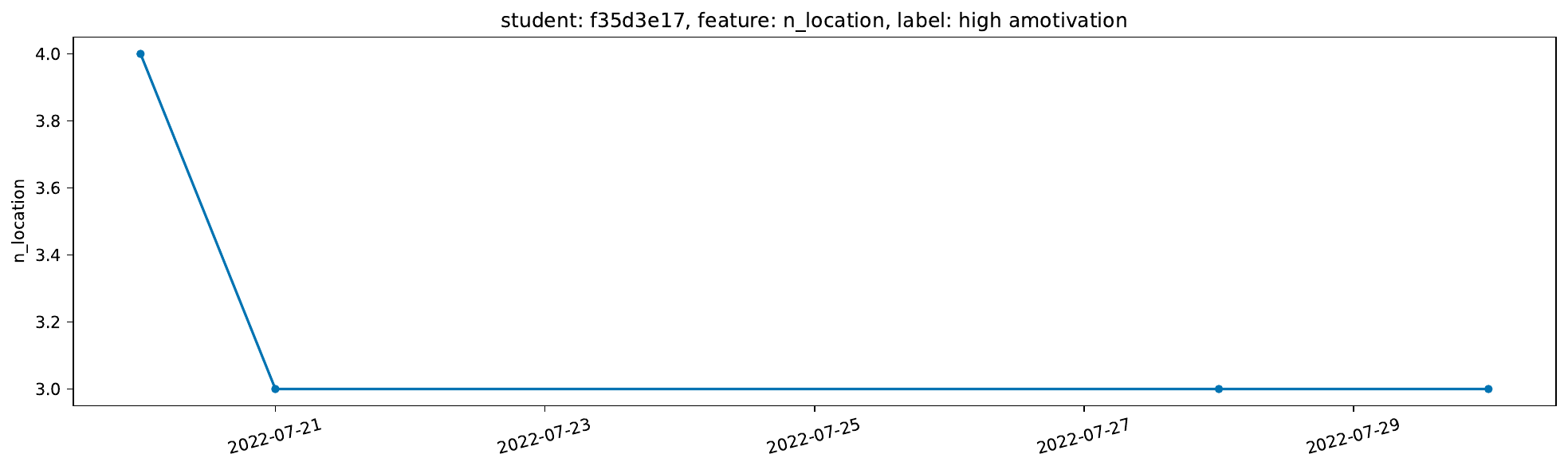}
\par\end{centering}
}
\par\end{centering}
\caption{\label{fig:Mobility-patterns}For an illustrative purpose, we show
\textbf{fake mobility patterns} of a student with \textquotedblleft low
amotivation\textquotedblright{} (a) and a student with \textquotedblleft high
amotivation\textquotedblright{} (b). The x-axis shows the date while
the y-axis shows the number of visited locations in one day.}
\selectlanguage{english}%
\end{figure*}

In our study, the numbers of travel days among students are often
different. In Figure \ref{fig:Mobility-patterns}, we show the mobility
patterns of two students. The student with ``low amotivation'' (a)
frequently travels to many places and has data for 15 days while the
student with ``high amotivation'' (b) visits fewer locations and
has data for only four days. Due to unequal length time series, we
cannot train ML classifiers to predict amotivation.

To address this problem, we impute missing data i.e. predict values
of $n_{location}$ and $dist_{travel}$ for missing days between time
series. There are many imputation methods for time series. Common
methods are ``linear'', ``nearest'', ``zero'', ``\foreignlanguage{english}{s-linear}'',
``previous'', and ``next'' \cite{kazijevs2023deep}. In our experiment,
we use ``linear'' -- the most popular method to impute missing
data.

\selectlanguage{english}%
Given a sequence of days $D=\{d_{1},...,d_{U}\}$, we construct a
vector $Z=\text{linspace}(0,U',U)$, where $U$ is the number of current
days, $U'$ is the number of imputed days, and $\text{linspace}()$
is a function to create linear points. For example, the student with
``high amotivation'' in Figure \ref{fig:Mobility-patterns}(b) has
$U=4$. Since we want to impute 15 days, $U'=15$. Note that the vector
$Z$ has a form $Z=[0,...,U']$ with a length of $U$. Next, we call\foreignlanguage{american}{
$F=[f_{1},...,f_{U}]$ as corresponding feature values of $D$ (here,
$F$ can be $n_{location}$ or $dist_{travel}$). Next, we fit a linear
function as $F=h(Z)$, and we construct a new }vector $Z'=\text{linspace}(0,U',U')$\foreignlanguage{american}{.
Note that the vector $Z'$ has a form $Z'=[0,...,U']$ with a length
of $U'$. Finally, we use the function $h()$ to predict feature values
for the new points $Z'$.}

\selectlanguage{american}%
At the end, each student has 15 features for $n_{location}$ and 15
features for $dist_{travel}$, which present values of $n_{location}$
and $dist_{travel}$ in 15 days. Note that as we impute the missing
data using a linear function, the variable $n_{location}$ now becomes
a continuous variable same as the variable $dist_{travel}$.

\subsubsection{Using \textit{trial\_id} as feature}

Given that the students were recruited in waves in 12 sequential mini
trials, the mini trial id can serve as a useful grouping variable
to group students who were enrolled at a similar point in time. For
example, \textit{trial\_id} 5 indicates enrolled dates between March
2022 and April 2022. There are 12 mini trials in total.

\begin{figure}[h]
\selectlanguage{english}%
\begin{centering}
\includegraphics[scale=0.3]{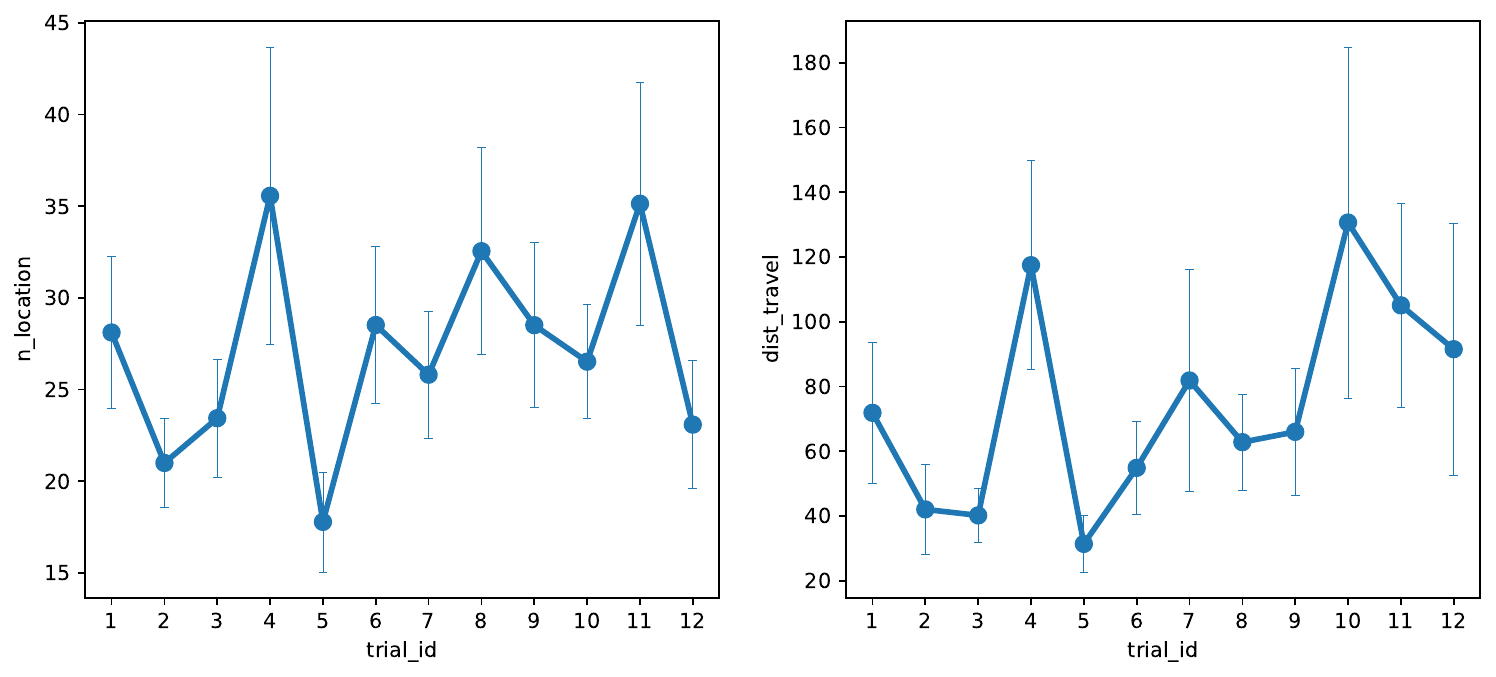}
\par\end{centering}
\caption{\label{fig:Distributions-of-two-variables}Distributions of variables
$n_{location}$ and $dist_{travel}$ in each trial. The y-axis shows
the mean value and its standard error.}
\selectlanguage{english}%
\end{figure}

For each \textit{trial\_id}, we show the mean and standard error of
$n_{location}$ and $dist_{travel}$ of all students in that trial
in Figure \ref{fig:Distributions-of-two-variables}. We observe that
our dataset contains heterogeneous samples, where students in some
trials have less moving patterns than the others e.g. \textit{trial\_id}
5.

Existing studies often conducted experiments in one university and
in one trial \cite{babic2017machine,schwan2021perceptions}. However,
some researches showed that ML classifiers worked well on one university
data but they failed to generalize to heterogeneous populations \cite{muller2021depression}.
In our study, the dataset consists of diverse students, collected
from 65 universities (or colleges) and 12 mini trials. We hypothesize
that \textit{trial\_id} can be a relevant feature to train ML classifiers
as it helps to distinguish sub-populations.

As shown in Table \ref{tab:The-final-tabular}, our final dataset
has 30 continuous variables and one nominal variable.

\begin{table}[h]
\selectlanguage{english}%
\caption{\label{tab:The-final-tabular}The final tabular format of our GPS
location dataset has 30 continuous variables and one nominal variable
\textit{trial\_id}.}

\begin{centering}
\begin{tabular}{|c|c|c|c|c|c|c|c|}
\hline 
\textbf{Student} & \multicolumn{3}{c|}{\textbf{$n_{location}$}} & \multicolumn{3}{c|}{\textbf{$dist_{travel}$}} & \textbf{\textit{trial\_id}}\tabularnewline
\hline 
 & day1 & ... & day15 & day1 & ... & day15 & \tabularnewline
\hline 
\hline 
\foreignlanguage{american}{abcxyz12} & 3.0 & ... & 3.3 & 2.8 & ... & 1.3 & 1\tabularnewline
\hline 
12abc456 & 2.0 & ... & 3.0 & 8.0 & ... & 4.3 & 3\tabularnewline
\hline 
\end{tabular}
\par\end{centering}
\selectlanguage{english}%
\end{table}

\subsection{Experimental settings}

In this section, we describe the list of baselines and classifiers
used in our experiments. We also explain how we choose the model hyper-parameters
and prepare training and test sets.

\subsubsection{Baselines}

For a comprehensive evaluation, we compare our oversampling method
SMOTE-PRED with 10 state-of-the-art baselines. They include SMOTE
\cite{chawla2002smote}, SMOTE-NC \cite{fernandez2018smote}, SMOTE
variants \cite{batista2003balancing,batista2004study,han2005borderline,nguyen2011borderline},
and two deep generative methods CTGAN \cite{Xu2019} and TVAE \cite{Xu2019,borisov2022deep}.
For reproducibility, we use the published source codes of the baselines
from the \textit{imbalanced-learn} library\footnote{\selectlanguage{english}%
https://imbalanced-learn.org/stable/\selectlanguage{english}%
} and \textit{SVD} package\footnote{\selectlanguage{english}%
https://github.com/sdv-dev/CTGAN\selectlanguage{english}%
}.

We also include AdaSyn \cite{he2008adasyn} and \textit{Imbalance}
-- the method keeps the original imbalanced dataset without oversampling.

\textbf{Hyper-parameters.} For the baselines, we use default hyper-parameters
as suggested in the library/package. For our method SMOTE-PRED, we
use an XGBoost \cite{chen2016xgboost} model as the predictor to re-correct
nominal values.

\subsubsection{Classifiers}

After rebalancing the training set, following other works \cite{babic2017machine,orji2022machine,orji2023modeling},
we train five popular ML classifiers to predict the student amotivation.
They include distance-based model \textit{k-nearest neighbors} (kNN),
kernel-based model \textit{support vector machine} (SVM), tree-based
model \textit{decision tree} (DT), and ensemble-based models \textit{random
forest} (RF) and \textit{XGBoost} (XGB). We tune their hyper-parameters
(see Table \ref{tab:Hyper-parameters-of-ML}) on the training set
by using ten-fold cross-validation. We report their results with the
best hyper-parameter set.

\begin{table}[h]
\selectlanguage{english}%
\caption{\label{tab:Hyper-parameters-of-ML}Hyper-parameters of ML classifiers
to predict student amotivation.}

\begin{centering}
\begin{tabular}{|c|V{\linewidth}|}
\hline 
\textbf{Classifier} & \textbf{Hyper-parameters}\tabularnewline
\hline 
\hline 
kNN & \textit{n\_neighbors}: \{1, 2, ..., 10\}\tabularnewline
\hline 
SVM & \textit{kernel}: \{linear, rbf\}

\textit{gamma}: \{0.1, 0.5, 1.0\}

\textit{C}: \{0.1, 0.5, 1.0, 10.0\}\tabularnewline
\hline 
DT & \textit{max\_depth}: \{1, 2, ..., 12\}\tabularnewline
\hline 
RF & \textit{n\_estimators}: \{1, 10, 50, 100\}

\textit{max\_depth}: \{1, 2, ..., 12\}

\textit{max\_features}: \{sqrt, log2, None\}\tabularnewline
\hline 
XGB & \textit{n\_estimators}: \{1, 10, 50, 100\}

\textit{max\_depth}: \{1, 2, ..., 12\}

\textit{max\_features}: \{sqrt, log2, None\}\tabularnewline
\hline 
\end{tabular}
\par\end{centering}
\selectlanguage{english}%
\end{table}

We repeat our experiments in \textit{ten times with random seeds}.
In each time, we construct training and test sets by randomly splitting
the dataset into 90\% for training and 10\% for testing. We run oversampling
methods combined with ML classifiers, and report the average \textit{area
under the curve} (AUC) score along with its standard deviation.

\subsection{Results and discussions}

Table \ref{tab:Main-experiments} reports AUC-scores of each oversampling
method combined with five ML classifiers.

\selectlanguage{english}%
\begin{table}[h]
\caption{\label{tab:Main-experiments}AUC $\pm$ (std) of each oversampling
method combined with five popular ML classifiers. \textbf{Bold} and
\uline{underline} indicate the best and second-best methods.}

\centering{}%
\begin{tabular}{|l|ccccccc|}
\hline 
\rowcolor{header_color}AUC-score & Imbalance & AdaSyn & SMOTE & SMOTE-NC & CTGAN & TVAE & SMOTE-PRED\tabularnewline
\hline 
\hline 
\multirow{2}{*}{kNN} & 0.5221 & 0.6065 & 0.6346 & \uline{0.6498} & 0.5382 & 0.5436 & \textbf{0.6735}\tabularnewline
 & (0.011) & (0.035) & (0.022) & (0.028) & (0.015) & (0.013) & (0.025)\tabularnewline
\hline 
\rowcolor{even_color}SVM & 0.4949 & 0.5854 & 0.5817 & 0.5572 & 0.4806 & \uline{0.5963} & \textbf{0.6198}\tabularnewline
\rowcolor{even_color} & (0.003) & (0.025) & (0.015) & (0.019) & (0.009) & (0.019) & (0.024)\tabularnewline
\hline 
DT & 0.5310 & \uline{0.5372} & 0.4992 & 0.5237 & 0.5167 & 0.5012 & \textbf{0.5491}\tabularnewline
 & (0.019) & (0.014) & (0.024) & (0.027) & (0.013) & (0.016) & (0.023)\tabularnewline
\hline 
\rowcolor{even_color}RF & 0.5147 & \uline{0.5246} & 0.5055 & 0.5173 & 0.5141 & 0.5113 & \textbf{0.5856}\tabularnewline
\rowcolor{even_color} & (0.018) & (0.016) & (0.023) & (0.027) & (0.013) & (0.020) & (0.025)\tabularnewline
\hline 
XGB & 0.5000 & \textbf{0.5695} & 0.5350 & 0.5335 & 0.5060 & 0.5214 & \uline{0.5600}\tabularnewline
 & (0.000) & (0.018) & (0.017) & (0.019) & (0.021) & (0.015) & (0.026)\tabularnewline
\hline 
\rowcolor{childheader_color}Average & 0.5125 & \uline{0.5646} & 0.5512 & 0.5563 & 0.5111 & 0.5348 & \textbf{0.5976}\tabularnewline
\hline 
\end{tabular}
\end{table}

\selectlanguage{american}%
Our method SMOTE-PRED are more accurate than other oversampling methods.
Across five ML classifiers, it is best performing with four classifiers
and second-best with another classifier. Its average improvement over
AdaSyn (the runner-up method) is \textasciitilde 3\% and SMOTE (the
most popular baseline) is \textasciitilde 4\%. More importantly,
it significantly outperforms Imbalance (the method without oversampling)
at \textasciitilde 8\%.

\selectlanguage{english}%
Most oversampling methods except CTGAN are better than Imbalance.
Interestingly, traditional methods like AdaSyn and SMOTE are more
effective than deep learning approaches like CTGAN and TVAE. This
can be explained by the fact that our dataset is quite small (only
hundreds of samples), which is not an ideal setting for deep learning.
SMOTE-NC -- an improved version of SMOTE to deal with nominal variables
is better than SMOTE in most cases. SMOTE-NC also becomes the third-best
method, following AdaSyn. However, the performance of SMOTE-NC is
still far away from ours, where the gap is \textasciitilde 4\%.

In summary, our method SMOTE-PRED is the best oversampling method,
and this result is consistent with many different ML classifiers.
As \textit{kNN is the most effective classifier} (it works well with
not only our method SMOTE-PRED but also AdaSyn, SMOTE, and SMOTE-NC),
\textit{we select it as the default classifier for our following experiments}.

\begin{figure}[h]
\begin{centering}
\includegraphics[scale=0.5]{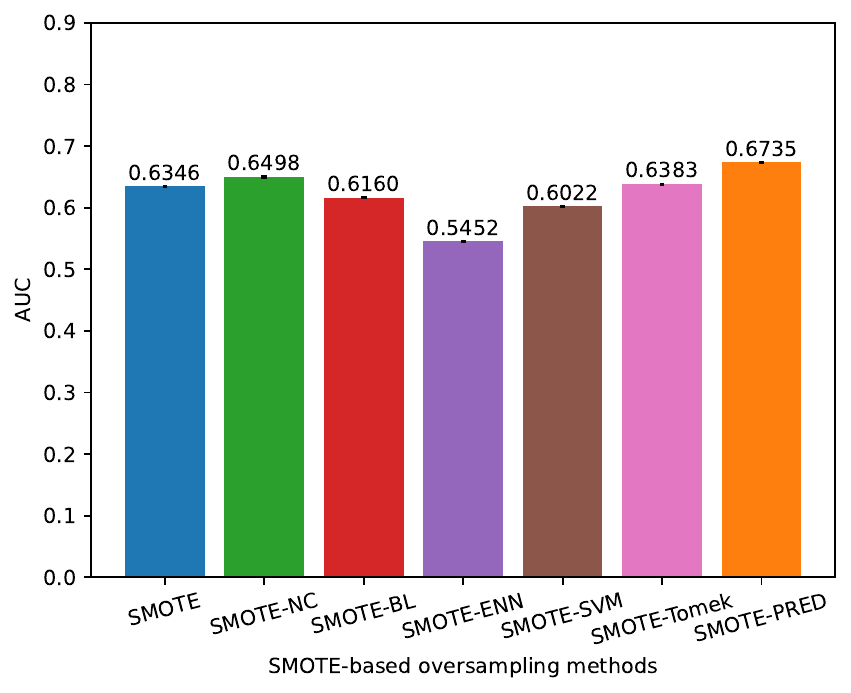}
\par\end{centering}
\caption{\label{fig:AUC-scores-of-SMOTE-variants}AUC-scores of our method
SMOTE-PRED and other SMOTE variants.}

\end{figure}

\textbf{Comparison with SMOTE variants.} Because our method is based
on SMOTE, we also compare it with SMOTE variants, including SMOTE-BL
\cite{han2005borderline}, SMOTE-ENN \cite{batista2004study}, SMOTE-SVM
\cite{nguyen2011borderline}, and SMOTE-Tomek \cite{batista2003balancing}.
Figure \ref{fig:AUC-scores-of-SMOTE-variants} shows the results.

Our method SMOTE-PRED has the best performance while SMOTE-NC is the
second-best method. SMOTE and SMOTE-Tomek behave similarly and they
become the third-best methods. Other SMOTE-based methods, especially
SMOTE-ENN, do not perform well on our dataset.

\selectlanguage{american}%

\subsection{Ablation studies}

We analyze our method under different configurations.

\subsubsection{Impact of imputed days}

As explained earlier, our dataset contains students who have unequal
length time series. In our experiments, we set the \textit{imputed\_days}
to 15 to generate equal length time series for each student. In this
study, we investigate the impact of the number of imputed days on
our method's performance.

\begin{figure}[h]
\selectlanguage{english}%
\begin{centering}
\includegraphics[scale=0.5]{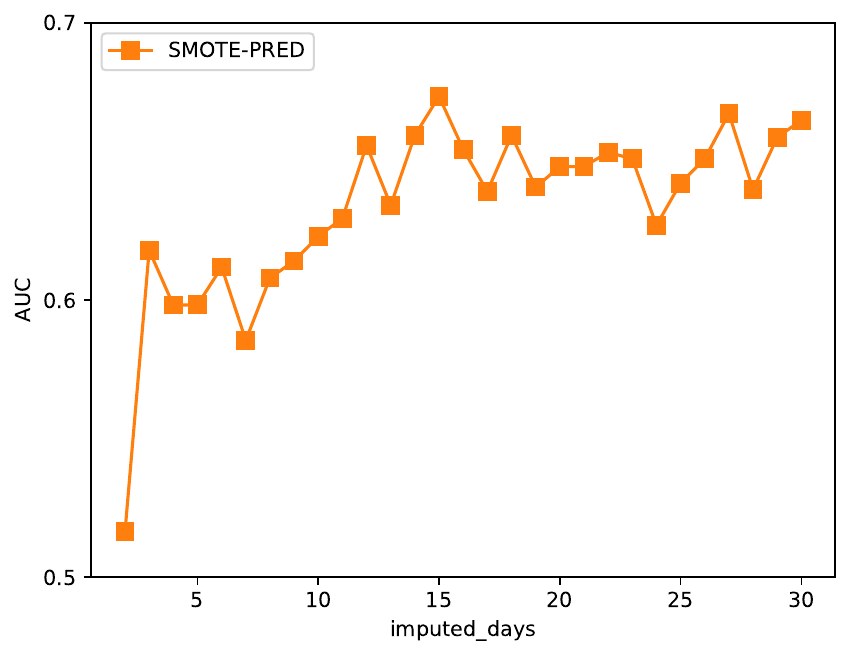}
\par\end{centering}
\caption{\label{fig:AUC-score-vs.-imputed-days}AUC-score vs. the number of
imputed days.}
\selectlanguage{english}%
\end{figure}

We adjust the value of \textit{imputed\_days} in the range of $[2,30]$,
and show our performance in Figure \ref{fig:AUC-score-vs.-imputed-days}.
Overall, our AUC-score increases when we set larger values for \textit{imputed\_days}.
When the value is large enough (i.e. \textit{imputed\_days} > 12),
our method always achieves AUC-score > 0.63.

\subsubsection{Impact of imputation method}

Recall that we leverage a ``linear'' function to impute missing
data. In this experiment, we investigate other imputation methods.
Table \ref{tab:AUC-score-vs.-imputation} shows our performance with
different imputation methods.

Two imputation methods ``linear'' and ``s-linear'' achieve the
best results. Other methods do not work well.

\begin{table}[h]
\selectlanguage{english}%
\caption{\label{tab:AUC-score-vs.-imputation}AUC-score vs. imputation methods.}

\begin{centering}
\begin{tabular}{|l|c|}
\hline 
\textbf{Imputation method} & \textbf{AUC}\tabularnewline
\hline 
\hline 
linear & \textbf{0.6735 (0.0246)}\tabularnewline
\hline 
nearest & 0.5976 (0.0259)\tabularnewline
\hline 
nearest-up & 0.5938 (0.0254)\tabularnewline
\hline 
zero & 0.5929 (0.0215)\tabularnewline
\hline 
s-linear & \textbf{0.6735 (0.0246)}\tabularnewline
\hline 
previous & 0.5929 (0.0215)\tabularnewline
\hline 
next & 0.5982 (0.0355)\tabularnewline
\hline 
\end{tabular}
\par\end{centering}
\selectlanguage{english}%
\end{table}

\subsubsection{Feature importance}

In this study, we examine feature importance (i.e. the relationship
between target variable and predictor variables). Recall that we only
extract and use three features (variables) in our experiments, including
the number of visited locations $n_{location}$, the average distance
traveled $dist_{travel}$, and \textit{trial\_id}.

First, we show how our AUC-score drops when we take out one-by-one
feature in Table \ref{tab:AUC-score-vs.-feature.}. Note that when
we remove one feature $n_{location}$ or $dist_{travel}$, it means
that we remove all 15 features related to $n_{location}$ or $dist_{travel}$
in Table \ref{tab:The-final-tabular}. We can see that $dist_{travel}$
is the most important feature, following by $n_{location}$ as they
cause the most drops in our AUC-scores (\textasciitilde 21\%) if
they are removed. Removing \textit{trial\_id} only leads to \textasciitilde 4\%
decrease in our AUC-score.

\begin{table}[h]
\selectlanguage{english}%
\caption{\label{tab:AUC-score-vs.-feature.}AUC-score vs. features. \textquotedblleft --
$n_{location}$\textquotedblright{} means that we remove all 15 features
related to $n_{location}$ (see Table \ref{tab:The-final-tabular}).}

\begin{centering}
\begin{tabular}{|l|c|}
\hline 
\textbf{Features} & \textbf{AUC}\tabularnewline
\hline 
\hline 
\{$n_{location}$, $dist_{travel}$, \textit{trial\_id}\} & 0.6735 (0.0246)\tabularnewline
\hline 
-- $n_{location}$ & 0.5836 (0.0203)\tabularnewline
\hline 
-- $dist_{travel}$ & 0.4697 (0.0275)\tabularnewline
\hline 
-- \textit{trial\_id} & 0.6360 (0.0155)\tabularnewline
\hline 
\end{tabular}
\par\end{centering}
\selectlanguage{english}%
\end{table}

Second, following \cite{misiuk2024multivariate}, we compute an \textit{importance
score} of each feature. We expect that removing an important feature
should have a large change in the classifier predictions whereas the
absence of an unimportant feature should have little effect. Given
a test set ${\cal D}_{test}=\{x_{j}\}_{j=1}^{N'}$, let $f(x_{j})$
and $f_{i}(x_{j})$ be the predictions of the original classifier
(i.e. using all features) and the modified classifier (i.e. removing
one feature $X_{i}$).

The \textit{importance score} of a feature $X_{i}$ is computed as:
\begin{equation}
s_{X_{i}}=\frac{\frac{1}{N'}\sum_{j=1}^{N'}\mid f(x_{j})-f_{i}(x_{j})\mid}{\frac{1}{N'}\sum_{j=1}^{N'}\mid f(x_{j})-\mu_{f}\mid},\label{eq:importance-score}
\end{equation}
where $N'$ is the number of samples in the test set ${\cal D}_{test}$
and $\mu_{f}=\frac{1}{N'}\sum_{j=1}^{N'}f(x_{j})$ is the mean value
of the predictions of the original classifier. Since the importance
score indicates the deviation from the original predictions, a higher
value for $s_{X_{i}}$ means the feature $X_{i}$ is more important.

\begin{figure}[h]
\selectlanguage{english}%
\begin{centering}
\includegraphics[scale=0.5]{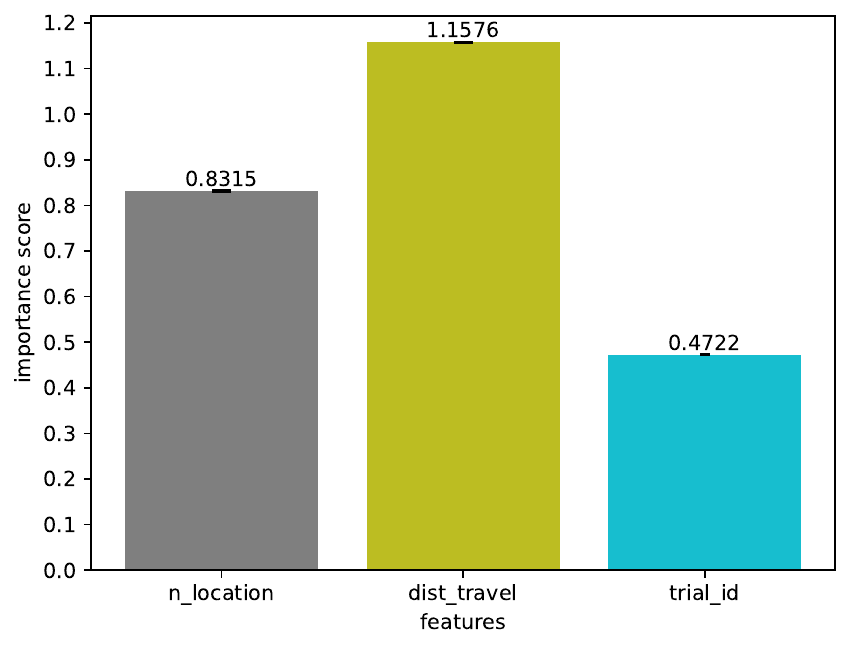}
\par\end{centering}
\caption{\label{fig:Importance-score}Feature importance. A higher score indicates
a more important feature.}
\selectlanguage{english}%
\end{figure}

We present the importance score of each feature in Figure \ref{fig:Importance-score}.
The feature $dist_{travel}$ has the highest score, indicating it
is the most important feature, following by $n_{location}$. The \textit{trial\_id}
is the least important feature. These results also agree with the
AUC-score drops in Table \ref{tab:AUC-score-vs.-feature.}.\selectlanguage{english}%

\section{Conclusion\label{sec:Conclusion}}

In this study, we demonstrate that GPS location data can effectively
differentiate between university students with ``low symptoms of
amotivation'' and those with ``high symptoms of amotivation''.
It offers valuable insights for the development of targeted mental
health interventions. Departing from previous approaches, we introduce
a novel and effective oversampling technique to rebalance imbalanced
training datasets in machine learning--based amotivation prediction.
Our method incorporates a crucial component to SMOTE, which predicts
nominal variables using a predictive model rather than an interpolation
step or choosing the most frequent nominal value.

We verify the effectiveness of our approach on a large-scale GPS location
dataset collected from Australian students, where it significantly
outperforms existing oversampling baselines.

\textbf{Future Work: }We plan to explore how the proposed component
can be integrated into other oversampling frameworks, including generative
models \cite{Xu2019} and large language models \cite{Nguyen2024},
to further enhance synthetic minority sample quality and robustness
in imbalanced learning tasks.

\section{Declarations}

\subsection*{Ethical Approval}

This study was approved by the University of New South Wales Human
Research Ethics Committee (Approval No: HC200466). ``Informed consent''
was obtained from all the participants and all methods were carried
out in accordance with relevant guidelines and regulations.

\subsection*{Funding}

This work was funded in part by a grant from the UK Wellcome Trust
(grant number: 303030/Z/23/Z).

\subsection*{Availability of data and materials}

To request access to de-identified data, please contact Professor
Jill Newby, via j.newby@unsw.edu.au.

\section*{Acknowledgment}

The Vibe Up Trial, from which the data analysed in this study were
obtained, was funded by the Medical Research Future Fund {[}MRFAI000028{]}.
The current research was partially supported by the Wellcome Trust
{[}303030/Z/23/Z{]}. AW was funded by a National Health and Medical
Research Council Investigator Grant {[}2017521{]}.

\balance

\bibliographystyle{plain}
\bibliography{reference}

\end{document}